\documentclass[preprint,12pt,authoryear]{elsarticle}

\usepackage{amssymb}
\usepackage{adjustbox}

\usepackage{amsmath}
\usepackage{tikz}
\usepackage{contour}
\usepackage{hyperref}
\contourlength{0.2pt} 
\usetikzlibrary{shapes.geometric, arrows.meta, positioning, fit, backgrounds, calc, decorations.pathreplacing, decorations.markings}
\usepackage{booktabs}

\begin{document}

\begin{frontmatter}



\title{GateTS: Versatile and Efficient Forecasting via Attention-Inspired routed Mixture-of-Experts} 



\author[lviv]{Kyrylo Yemets} 
\author[eindhoven]{Mykola Lukashchuk}
\author[lviv2,london]{Ivan Izonin}

\affiliation[lviv]{organization={Department of Artificial Intelligence, 
  Lviv Polytechnic National University},
            addressline={kyrylo.v.yemets@lpnu.ua}, 
            city={Lviv},
            postcode={79005}, 
            country={Ukraine}}

\affiliation[eindhoven]{organization={Department of Electrical Engineering, 
  Eindhoven University of Technology},
            addressline={m.lukashchuk@tue.nl}, 
            city={Eindhoven},
            postcode={5612 AE}, 
            country={Netherlands}}

\affiliation[lviv2]{organization={Department of Artificial Intelligence, 
  Lviv Polytechnic National University},
            addressline={ivan.v.izonin@lpnu.ua}, 
            city={Lviv},
            postcode={79005}, 
            country={Ukraine}}

\affiliation[london]{organization={The Bartlett School of Sustainable Construction, 
  University College London},
            addressline={ivanizonin@gmail.com}, 
            city={London},
            postcode={WC1E 7HB}, 
            country={United Kingdom}}

\begin{abstract}
Accurate univariate forecasting remains a pressing need in real-world systems—such as energy markets, hydrology, retail demand, and IoT monitoring — where signals are often intermittent and horizons span both short- and long-term. While transformers and Mixture-of-Experts (MoE) architectures are increasingly favored for time-series forecasting, a key gap persists: MoE models typically require complicated training with both the main forecasting loss and auxiliary load-balancing losses, along with careful routing/temperature tuning, which hinders practical adoption. In this paper, we propose a model architecture that simplifies the training process for univariate time series forecasting and effectively addresses both long- and short-term horizons, including intermittent patterns. Our approach combines sparse MoE computation with a novel attention-inspired gating mechanism that replaces the traditional one-layer softmax router. Through extensive empirical evaluation, we demonstrate that our gating design naturally promotes balanced expert utilization and achieves superior predictive accuracy without requiring the auxiliary load-balancing losses typically used in classical MoE implementations. The model achieves better performance while utilizing only a fraction of the parameters required by state-of-the-art transformer models, such as PatchTST. Furthermore, experiments across diverse datasets confirm that our MoE architecture with the proposed gating mechanism is more computationally efficient than LSTM for both long- and short-term forecasting, enabling cost-effective inference. These results highlight the potential of our approach for practical time-series forecasting applications where both accuracy and computational efficiency are critical.
\end{abstract}




\end{frontmatter}



\section{Introduction}

Time series forecasting is pivotal across various domains, including finance, energy, healthcare, and logistics. Accurate predictions of future values based on historical data enable informed decision-making and strategic planning. Traditional forecasting models, such as ARIMA and exponential smoothing, often struggle with capturing complex patterns and dependencies inherent in time series data, especially when dealing with long-term forecasts or limited data availability. Advancements in deep learning have introduced models like Transformers and attention mechanisms, which have shown promise in handling sequential data. These models excel in capturing long-range dependencies and have been adapted for time series forecasting tasks. However, they often require large amounts of data and are primarily designed for multivariate time series, limiting their effectiveness on univariate datasets or scenarios with scarce data.

The Mixture-of-Experts (MoE) architecture, originally developed for large-scale language models, offers a compelling approach to address these challenges. A Mixture-of-Experts architecture comprises two principal components: the experts, which are specialized sub-models trained to handle distinct aspects of the input space, and the gating (or routing) mechanism, which selects the most relevant experts for each input. The gating network’s decisions determine both the computational path and the specialization of the experts. By combining multiple specialized models (experts) and dynamically selecting the most relevant ones for a given input, MoE architectures can capture diverse patterns and improve model capacity without significantly increasing computational complexity.  This is achieved through a specialized training procedure incorporating an additional auxiliary loss, which facilitates balanced information distribution across experts. However, this approach may introduce a trade-off, potentially reducing predictive accuracy.

Building upon these insights, we propose a model architecture: MOE Time Series Forecasting with Attention-Inspired Gating Mechanism. That unites the conditional-capacity benefits of Mixture-of-Experts with a router expressly crafted for temporal data. Classic one-layer gates are appealingly simple but tend to activate only a small subset of experts unless guided by auxiliary load-balancing losses; without such regularization, most experts remain dormant, preventing the network from realizing its full capacity and leaving accuracy on the table. In contrast, our attention-inspired gate learns to distribute information across experts through the same end-to-end forecasting loss, dynamically re-weighting their contributions according to the input sequence. This self-adaptive routing allows the model to capture diverse temporal patterns and scales, making it especially effective for univariate series and for settings where data are scarce and traditional methods struggle.

In this paper, the authors developed a novel time-series forecasting model called GateTS. The novelty of GateTS stems from a new gating mechanism that enables the use of MoE architecture for time-series forecasting. GateTS demonstrates that time-series forecasting can be effectively approached through Mixture-of-Experts architectures when equipped with our proposed specialized gating mechanism, which improves upon classical MoE. Our main improvement stems from enhancing standard MoE routing by utilizing experts more effectively. While conventional MoE gating fails to properly route temporal patterns to appropriate experts, our mechanism enables expert specialization across different forecasting regimes, resulting in a model that can maintain state-of-the-art performance across both short and long-term horizons without the computational burden of fully activated expert networks during inference time.


Generally, the main scientific contributions are:
\begin{itemize}
    \item We introduce an attention-inspired gating mechanism that replaces the standard one-layer softmax router in sparse MoE and enables stable training \emph{without} any auxiliary load-balancing loss (i.e., the “auxiliary loss” of \citep{gshard}), while maintaining balanced expert utilization; see \ref{subsec:gating}.
    \item We present \textit{GateTS}, a sparse MoE architecture tailored to univariate time-series forecasting that jointly addresses short-term and long-term horizons, including intermittent signals, and achieves leading predictive accuracy with substantially fewer parameters and lower inference cost than PatchTST and LSTM baselines; see \ref{sec:moe_arch}.
\end{itemize}


\section{Background}

\subsection{Time series forecasting}

Recent Transformer research for time-series forecasting has largely focused on long-horizon multivariate tasks. Architectures such as Autoformer \citep{autoformer}, Informer \citep{informer}, iTransformer \citep{itransformer}, and PatchTST \citep{patchtst} reduce the quadratic cost of vanilla self-attention while introducing strong inductive biases—e.g., trend–seasonality decomposition, probabilistic sparsification, inverted attention, and temporal patching—to predict dozens or hundreds of future steps. Follow-up models, including FEDformer \citep{fedformer}, TimesNet \citep{timesnet}, and ETSformer \citep{etsformer}, further develop frequency-domain and exponential-smoothing perspectives and currently dominate long-term multivariate leaderboards. 

Autoformer decomposes each series into slowly varying trend and seasonal components and replaces dot-product attention with an auto-correlation mechanism that re-uses historical lags \citep{autoformer}. Informer adopts ProbSparse attention, sampling only the highest-magnitude query–key pairs, and combines this with layer-wise distillation to extend receptive fields to thousands of time steps at sub-quadratic cost \citep{informer}. iTransformer inverts the usual attention order by treating time steps as channels; this inversion highlights frequency structure and yields competitive multivariate results with fewer parameters \citep{itransformer}. PatchTST concatenates non-overlapping temporal patches before feeding them to a vanilla encoder-only Transformer, striking a compromise between locality and global context while preserving the standard $O(L^2)$ attention inside each patch \citep{patchtst}. FEDformer augments the encoder–decoder scheme with frequency-domain filtering and learns a structured low-rank approximation that scales linearly with sequence length \citep{fedformer}. TimesNet formulates a two-dimensional variation block that unifies forecasting, imputation, classification, and anomaly detection within a single spectral backbone \citep{timesnet}. ETSformer integrates exponential-smoothing attention and modular decomposition layers to learn level, growth, and seasonality directly in the Transformer, improving both interpretability and efficiency \citep{etsformer}.

However, this prevailing orientation has left the one-step-ahead univariate setting comparatively under-explored. The sophisticated architectural components optimized for capturing complex cross-channel dependencies and long-range patterns can become over-parameterized and statistically mismatched when the forecast horizon shrinks to a single point and the input is a single series. Concurrently, a complementary path to model scaling has been revitalized through the Mixture-of-Experts (MoE) paradigm, which enables sparse conditional computation. Initially developed in systems like GShard \citep{gshard} and Switch Transformer \citep{switch}, MoE has been successfully adapted for temporal data in models such as the 2.4-billion-parameter TimeMoE \citep{timemoe}, demonstrating its viability for time-series analysis.

Short-term forecasting is still central to inventory control, trading, and streaming anomaly detection \citep{onestep}.  Yet Transformer variants tuned for multivariate, 96- to 1024-step tasks routinely employ dimensions an order of magnitude larger than classical autoregressive models, making them prone to over-fitting when the target series is short or noisy. Empirical studies often exclude horizons of one or four steps, leaving a methodological gap for architectures explicitly tailored to univariate, short-term dynamics. Therefore, the objective of this study is to develop an efficient and versatile architecture capable of accurately capturing both short- and long-term temporal dependencies for forecasting tasks.

\subsection{Gating mechanism}

Consider an input tensor $\mathbf{x}\in\mathbb{R}^{T\times d}$ (e.g., a sequence of T tokens each embedded in d dimensions).
Let $\{f_e(\,\cdot\,)\}{e=1}^{E}$ be a set of E expert functions (“experts”), each producing an output in $\mathbb{R}^{m}$.
The gating network $G{\boldsymbol\theta}$ is a parametric map $G_{\boldsymbol\theta}:\;\mathbb{R}^{T\times d}\;\longrightarrow\;\Delta^{E-1},$ where $\Delta^{E-1}=\{\,\mathbf{p}\in\mathbb{R}^{E}\mid p_e\ge 0,\;\sum_{e=1}^{E}p_e=1\,\}$ is the (E-1)-simplex.
For an input $\mathbf{x},$ the gate produces a routing distribution
\begin{equation}
\mathbf{p}(\mathbf{x}) = \bigl(p_1(\mathbf{x}),\ldots,p_E(\mathbf{x})\bigr)=G_{\boldsymbol\theta}(\mathbf{x}),    
\end{equation}
 which assigns a non-negative mixing weight to every expert.

Sparse MoE routing has re-emerged as a dominant scaling strategy in language modelling. Mixtral exemplifies this trend by selecting two out of eight expert feed-forward blocks per token, thereby expanding representational capacity without linear cost growth \citep{mixtral}. The foundational GShard project introduced an auxiliary load-balancing loss and random routing to ensure gradient flow to all experts during training \citep{gshard}, while Switch Transformer refined this idea with a single-expert router, adaptive capacity constraints, and stable mixed-precision optimisation \citep{switch}. Time-series research is beginning to adopt the same principles. The TimeMoE family activates only a subset of experts for each timestamp, allowing a 2.4-billion-parameter decoder-only model to operate with dense-model inference cost and to deliver zero-shot accuracy across arbitrary horizons and datasets \citep{timemoe}.

A fully connected router followed by a soft-max remains the canonical gating mechanism in Mixture-of-Experts (MoE) layers, providing a convex combination of expert outputs conditioned on the input representation.  The importance of this gate for scalable conditional computation was first highlighted in large-scale deployments such as GShard, which showed that, without an explicit load-balancing objective, the router collapses onto a small subset of experts and negates the intended capacity gains \citep{gshard}.  To counter expert collapse, Noisy Top-k routing perturbs the pre-softmax logits with Gaussian noise, encouraging exploration and leading to more uniform expert utilisation during training \citep{noisegating}.

\section{Method}\label{sec:method}

Time-series forecasting seeks a mapping $f: \mathbb{R}^{T}\!\rightarrow\!\mathbb{R}^{H}$ that predicts the next $H$ future values of a signal given its past $T$ context observations.
Let $x_{1:T}=\{x_1,\ldots,x_T\}$ denote the known history and $y_{T+1:T+H}=\{y_{T+1},\ldots,y_{T+H}\}$ the future horizon.
The goal is to minimise a loss $\mathcal{L}(y,\hat{y})$ between forecasts $\hat{y}=f(x_{1:T})$ and ground truth $y_{T+1:T+H}$. 

\subsection{MOE architecture}\label{sec:moe_arch}

\begin{figure}[ht]
\centering
\adjustbox{width=\textwidth,center}{\begin{tikzpicture}[
  font=\normalsize, line width=1pt,
  router/.style={draw, fill=cyan!30, rounded corners, minimum width=18mm, minimum height=8mm},
  expertSel/.style={draw, fill=green!20, rounded corners, minimum width=22mm, minimum height=8mm},
  expertDrop/.style={draw, fill=red!20, rounded corners, minimum width=22mm, minimum height=8mm},
  merge/.style={circle, draw, minimum size=6mm},
  arrow/.style={-{Latex[length=2.5mm]}, thick},
  timeline/.style={draw, ->, thick},
  sample/.style={circle, draw, fill=black, inner sep=1pt},
  context/.style={draw=none, fill=orange!50, opacity=0.3, inner sep=0pt},
  probvec/.style={draw, rounded corners, minimum width=14mm, minimum height=10mm, align=center},
  inputblock/.style={draw, fill=purple!30, rounded corners, minimum width=8mm, minimum height=8mm},
  outputblock/.style={draw, fill=orange!30, rounded corners, minimum width=8mm, minimum height=8mm},
  LinearHead/.style={draw, fill=orange!25, rounded corners, minimum width=22mm, minimum height=8mm},
  LinearHeadIn/.style={draw, fill=purple!25, rounded corners, minimum width=22mm, minimum height=8mm},
  ProbBlockSelected/.style={draw, fill=green!30, rounded corners, minimum width=8mm, minimum height=8mm},
  ProbBlockNotSelected/.style={draw, fill=red!30, rounded corners, minimum width=8mm, minimum height=8mm},
  PreprocIn/.style={draw, fill=purple!25, rounded corners, minimum width=16mm, minimum height=8mm},
]
\draw[->] (1,2) -- (11,2);

\node[draw, fill=purple!25, text=blue, rounded corners, minimum width=20mm, minimum height=6mm] (context) at (3.2, 3.4) {Context Window $\mathbb{R}^{T}$};

\foreach \i in {0,...,4}{
  \node[inputblock] (cube\i) at (2 + 0.9*\i, 2.4) {};
}

\node[draw, rounded corners=1mm, thick, fit=(cube0) (cube4), inner sep=6pt] (vectorin) {};

\node[draw, fill=orange!25, text=blue, rounded corners, minimum width=20mm, minimum height=6mm] (context) at (8.4, 3.4) {Forecasting Horizon $\mathbb{R}^{H}$};

\foreach \i in {0,...,3}{
  \node[outputblock] (сubeout\i) at (7.2 + 0.9*\i, 2.4) {};
}

\node[draw, rounded corners=1mm, thick, fit=(сubeout0) (сubeout3), inner sep=6pt] (vectorout) {};

\node[expertSel] (expert1) at (6.2, -0.7) {Expert $E_1$};
\node[expertDrop] (expert13) at (6.2,-1.6) {Expert $E_{2}$};
\node[expertDrop] (expertE) at (6.2,-2.5) {Expert $E_{e}$};

\node[draw=black, dashed, rounded corners=2mm, fit=(expert1) (expert13) (expertE), inner sep=6pt] (expertBox) {};
\node[above=1pt of expertBox] {Expert Bank};

\node[merge] (sum) at (9.2,-1.6) {$\Sigma$};
\node[ProbBlockSelected] (prob1) at (8.2, -4.2) {$0.98$};
\node[ProbBlockNotSelected] (prob2) at (9.2, -4.2) {$0.01$};
\node[ProbBlockNotSelected] (prob3) at (10.2, -4.2) {$0.01$};
\node[draw, rounded corners=1mm, thick, fit=(prob1) (prob2) (prob3), inner sep=6pt] (vectorprob) {};

\node[LinearHeadIn] (topk) at (3.2, -4.2) {Gating};
\node[PreprocIn] (preproc) at (3.2, -1.6) {Prepare};

\draw[arrow] (preproc) -- (expert1);
\draw[arrow] (preproc) -- (expert13);
\draw[arrow] (preproc) -- (expertE);
\draw[arrow] (preproc) -- (topk);
\draw[arrow] (topk) -- (vectorprob);
\draw[arrow] (prob1) --  (sum);
\draw[arrow, dotted] (prob2) -- (sum);
\draw[arrow, dotted] (prob3) -- (sum);

\draw[arrow] (expert1) --  (sum);
\draw[arrow, dotted] (expert13) -- (sum);
\draw[arrow, dotted] (expertE) -- (sum);

\node[LinearHead] (resultinghead) at (12.2, -1.6) {embedding $\mathbb{R}^{d}$};
\draw[arrow] (sum) -- (resultinghead);

\draw[orange, ultra thick, ->, >=stealth] (resultinghead.north east) -- (vectorout.north east);
\draw[orange, ultra thick, ->, >=stealth] (resultinghead.south west) -- (vectorout.south west);

\coordinate (midpoint) at ($(resultinghead)!0.5!(vectorout)$);

\node[text=orange, font=\bfseries] at (midpoint) {Linear projection};

\node[LinearHeadIn] (inputhead) at (-0.3, -1.6) {embedding $\mathbb{R}^{d}$};
\draw[arrow] (inputhead) -- (preproc);

\draw[purple, ultra thick, ->, >=stealth] (vectorin.south east) -- (inputhead.south east);
\draw[purple, ultra thick, ->, >=stealth] (vectorin.north west) -- (inputhead.north west);

\coordinate (midpointin) at ($(inputhead)!0.5!(vectorin)$);
\node[font=\bfseries, text=purple] at (midpointin) {Linear projection};

\end{tikzpicture}}
\caption{ Architecture of mixture of experts model adapted for time series forecasting, with choosing top-k best experts for inference task. Where $T$ is context size, and $H$ forecasting horizon.
}
\label{fig:gateTS-architecture}
\end{figure}
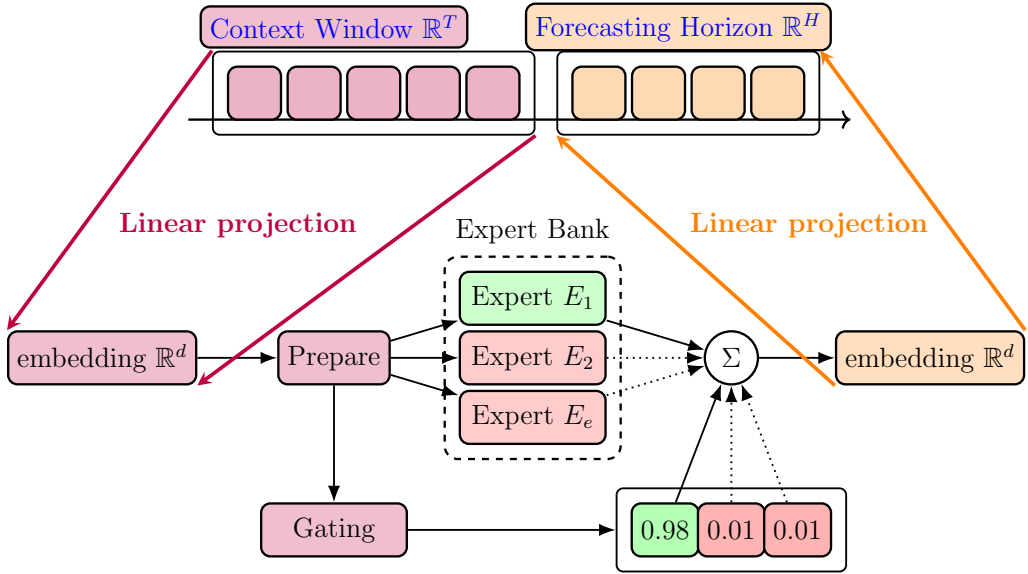

The model accepts an input sequence $x_{1:T}$ that is first linearly projected into a $d_{\text{model}}$ -dimensional token space and then augmented with a learnable positional embedding so that temporal order is retained despite the permutation-invariant nature of self-attention.
A shallow decoder pre-block applies multi-head self-attention, dropout, and layer normalisation, allowing the representation to aggregate information across all time-steps while remaining stable during optimisation. This block is called the Prepare Block. The whole architecture is shown in Figure \ref{fig:gateTS-architecture}. Gating mechanism will be described later in section \ref{subsec:gating}.

Given hidden representation $H \in \mathbb{R}^{T\times d_{\text{model}}}$, a learned router produces logits $r \in \mathbb{R}^{T\times N_{\text{exp}}}$ over $N_{\text{exp}}$ experts.
The Top--k operator keeps only the largest $k$ logits per token, yielding mask $m \in \{0,1\}^{T\times N_{\text{exp}}}$.
The Top–k operator is implemented token-wise: for every time-step $t\!\in\![1,T]$ the router first extracts the index set of the k largest logits as shown in the following equation:
\begin{equation}
    \mathcal{S}{k}^{(t)}=\operatorname*{TopK}{k}\!\bigl(r_{t,\:1:N_{\text{exp}}}\bigr),
\end{equation}
then forms a binary mask
$m_{t,e}=\mathbf{1}\!\bigl[e\in\mathcal{S}{k}^{(t)}\bigr],\qquad
m\in\{0,1\}^{T\times N{\text{exp}}},$
and sparsifies the logits via element-wise multiplication $r\leftarrow r\odot m$.
Finally, the surviving k entries are renormalised:
\begin{equation}
{p}_t=\frac{{r}_t\odot {m}_t}{{1}^\top({r}_t\odot {m}_t)}.
\end{equation}
yielding an exactly k-sparse routing distribution $p\in[0,1]^{T\times N_{\text{exp}}}$ whose non–zero probabilities preserve the relative ordering of the original logits.
Routing probabilities are then resulting in an \emph{exactly} $k$--sparse categorical distribution that back–propagates through the soft--max while zeroing out unselected experts.
Each active expert consists of its own self--attention sub--layer, layer normalisation, dropout, and position--wise feed--forward module.

Outputs of the selected experts are combined through a weighted sum using the gating probabilities, followed by an additional dropout and layer normalization step to ensure regularization and numerical stability.

A final linear head maps the aggregated hidden states to the target forecasting dimension, supporting both single-step and multi-step prediction tasks. 

 
\subsection{Attention-inspired Gating}\label{subsec:gating}

The Attention-Inspired gating converts every token representation into a probability distribution over E experts by running a one-head, query–key attention where the queries are the experts themselves.

Each token vector is linearly projected to create keys. This mirrors the key step in Transformer self-attention and provides a feature space in which similarity to expert queries will be measured.

\begin{equation} \label{eq:gating_key_proj}
K \;=\; W_k\,x + b_k,
\qquad W_k\in\mathbb{R}^{d\times d},\;     
\end{equation} where $d$ is the dimension of embeddings. 

EQ is a learnable, input-agnostic parameter matrix (i.e., it is not derived from the token embeddings). The same query matrix is broadcast to all batch elements and time steps.

\begin{equation}\label{eq:queries}
\text{EQ} \;=\; W_q\ + b_q,
\qquad W_q\in\mathbb{R}^{d\times e}.\;
\end{equation}

Instead of traditional scaled dot-product similarity applied to keys and queries, we used the Kronecker outer product. As shown in  \citep{attentionSymmetry} keys and queries can become correlated, so the Kronecker product was used to capture different features in keys and queries. This lifts similarity from a linear match in a shared subspace to a second-order interaction space that encodes all cross-dimension terms between queries and keys.

\begin{equation}\label{eq:kron_feat}
Z
=\;
K\;\otimes\;\text{EQ},  \quad Z \in\mathbb{R}^{d^{2}\times e},
\end{equation}

Then we project the Kronecker product into expert space. And scaling by  \(\alpha=d^{-\tfrac12}\) prevents the inner products from growing with dimension, a method first introduced in \citep{allyouneed} and later analysed for numerical stability in large-scale MoEs \citep{dotproduct}.

\begin{equation}\label{eq:kron_score}
S
=\;
\frac{W_{e}^{\top} Z}{\alpha},
\qquad
\alpha=d^{-\tfrac12},
\quad
W_e \in\mathbb{R}^{ d^{2} \times e}
\end{equation}

Where $W_e$ is the projection from $d^2$ space into $e$.

For each token, the vector is normalised:

\begin{equation}\label{eq:softmax}
p_{b,t,e}
\;=\;
\frac{\exp\bigl(S_{b,t,e}\bigr)}
{\sum_{e{\prime}=1}^{E}\exp\bigl(S_{b,t,e{\prime}}\bigr)}. 
\end{equation}

\begin{figure}[ht]\label{fig:gating}
\centering
\adjustbox{width=\textwidth,center}{\begin{tikzpicture}[
    node distance=1.5cm and 3.0cm,
    var/.style={
        rectangle,
        draw,
        thick,
        text centered,
        minimum size=1.5cm,
        rounded corners,      
        fill=orange!30,       
        text=blue             
    },
    arrow/.style={-Stealth, thick},
    eq_label/.style={midway, above, font=\small, text centered}
]

\node[var] (x) {$x$};
\node[var, right=of x] (K) {$K$};
\node[var, above=of K] (EQ) {$EQ$};
\node[var, right=of K] (Z) {$Z$};
\node[var, right=of Z] (S) {$S$};
\node[var, right=of S] (p) {$p$};


\node (q_source) [left=3cm of EQ] {};
\draw[arrow, dashed] (q_source) -- node[eq_label] {$W_q + b_q$~\eqref{eq:queries}} (EQ);

\draw[arrow] (x) -- node[eq_label] {$W_k x + b_k$~\eqref{eq:gating_key_proj}} (K);
\draw[arrow] (K) -- node[eq_label] {$K \otimes EQ$~\eqref{eq:kron_feat}} (Z);
\draw[arrow] (EQ) -- (Z); 
\draw[arrow] (Z) -- node[eq_label] {$\frac{W_e^T Z}{\alpha}$ \eqref{eq:kron_score}} (S);
\draw[arrow] (S) -- node[eq_label] {$\text{Softmax}(S)$ \eqref{eq:softmax}} (p);

\end{tikzpicture}}
\caption{Information flow through the attention-inspired gating mechanism, showing the transformation from input $x$ to expert routing probabilities $p$ via key projection \eqref{eq:gating_key_proj}, the Kronecker product with expert queries \eqref{eq:kron_feat}, scoring \eqref{eq:kron_score}, and the softmax normalisation \eqref{eq:softmax}.}
\end{figure}
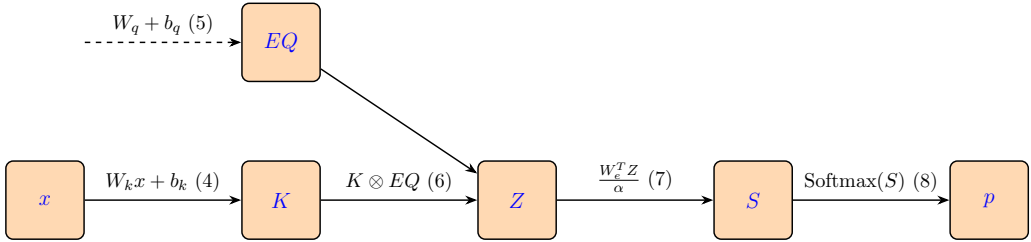

The resulting tensor gives gating probabilities that sum to 1 across experts. 
The complete gating procedure is illustrated in Figure~\ref{fig:gating}.

\subsection{Training Strategy} \label{sec:training}

The entire network is trained end--to--end with a single mean--squared--error loss
$\mathcal{L}_{\text{MSE}}=\|y_{T+1:T+H}-\hat{y}\|_2^2$.
\emph{No} auxiliary entropy or load--balancing losses are used. The  model's parameters $\boldsymbol\theta$  are updated through the gradients that come from the forecasting objective: 
\begin{equation}
\boldsymbol\theta
\leftarrow
\boldsymbol\theta
-\eta\,\frac{1}{B}\sum_{b=1}^{B}\nabla_{\boldsymbol\theta}\mathcal{L}^{(b)}_{\text{MSE}},    \end{equation}

where $\eta$ is the learning-rate, and $B$ is the batch size.

This minimalist design removes hyper-parameters related to balancing coefficients and keeps the optimisation focus on predictive accuracy, while this can lead to some experts not being used in forecasting at all, or becoming experts for anomalies. 

\section{Related Work}

Beyond pure neural architectures, related ideas appear in probabilistic models: recurrent switching dynamical systems use a learned softmax gate to select among multiple linear dynamical experts at each time step, yielding an HMM-like decomposition of complex trajectories \citep{recurent_gating}.  This probabilistic gate is conceptually similar to MoE routing and motivates our reuse of the mechanism inside neural networks; we refer to the resulting variant as HMM gating.

Recurrent Switching Linear Dynamical Systems (rSLDS) proposed in \citep{recurent_gating} using the gating (or transition) distribution as a covariate-dependent categorical model: the log-odds of moving into discrete state e are a linear function of the current continuous covariates, regularised with Pólya–Gamma augmentation so that Bayesian inference remains conjugate.  We re-engineer this statistical gate for modern neural Mixture-of-Experts (MoE) by (i) replacing the linear predictor with an attention-style dot-product between covariate embeddings and learnable expert keys, and (ii) retaining the rSLDS “Markov” bias as a state-prior term.  The result preserves the probabilistic semantics of the original gate while ensuring differentiability and GPU efficiency required by large-scale neural networks.

\subsection{HMM Gating}

This gating has some similarities with our gating \ref{subsec:gating} and in this section, the main differences between them will be described. 

This step embeds raw covariates into the same space as the expert keys, mirroring the key-projection in scaled dot-product attention but in a different dimension in comparison to \eqref{eq:gating_key_proj} we have the following relation:
\begin{equation}
Z \;=\; X W ,\qquad Z\in\mathbb{R}^{T\times d},     
\end{equation}

where $T$ - dimension of input, and $d$ - dimension of embeddings.
In this variant, Q is treated as a learnable matrix as in \eqref{eq:queries}; subsequently, we replace the Kronecker product in \eqref{eq:kron_score} with a dot product:
\begin{equation}
S \;=\; \frac{1}{\sqrt{d}}Z^{\top} Q,     
\end{equation}

The vector $\mathbf{m}$ is a learnable vector, that plays the role of a log-prior over experts, echoing the latent-state transitions used in Hidden-Markov mixtures of experts and enabling regime-dependent preferences that are independent of the current covariates.

\begin{equation}
\psi_{e} \;=\; S \;+\; m_e,     
\end{equation}

The soft-max transforms the logits into a proper categorical distribution. Where we have the probability of each expert as in \eqref{eq:softmax}. 
Overall, the main difference between the two gating mechanisms is in the Kronecker product  and dot product in HMM gating  and summarization after that with m vector.

\section{Results}\label{sec:results}

In this section, we evaluate three Mixture-of-Experts (MoE) variants: GateTS, which employs an attention-inspired gating mechanism; MOEhmm, which uses an HMM-based router; and MOEclassic, an ablation with a single linear gating layer to isolate the effect of the routing mechanism. As non-MoE baselines, we include a univariate LSTM \citep{lstm} and PatchTST \citep{patchtst} as a strong state-of-the-art Transformer baseline for forecasting.


Models trained with AdamW ($\beta_1{=}0.9,\,\beta_2{=}0.999,\,\text{weight decay}{=}1\times10^{-2}$), learning rate $1\times10^{-3}$ and a cosine schedule with warm-up. Data splits are 80\% for training and 10\% for validation and test datasets. Batch size is 256. Experiments were conducted on M1 Pro with 16GB of RAM for about 20 hours. 

\subsection{Datasets}

The sunspot dataset \citep{sunspot} is a daily record of the international sunspot number extending from 8 January 1818 through 31 May 2020 (73 924 data points). It provides a classic benchmark for non-stationary, low-frequency cyclical behavior. The original series contained occasional missing values; the Last Observation Carried Forward (LOCF) method was applied to ensure continuity before normalization and model input. This long historical span, spanning over two centuries, makes the sunspot series suitable for testing the ability of forecasting models to capture slow-moving trends, multi-scale seasonality, and occasional outliers.

The Saugeen River dataset \citep{saugeen} comprises daily mean flow measurements at Walkerton, Ontario, from 1 January 1915 to 31 December 1979, for a total of 23 741 observations. This dataset exhibits pronounced seasonal cycles tied to annual hydrological patterns, inter-annual variability, and occasional extreme-flow events. No missing values are present.

The wind-power series \citep{wind} records instantaneous power output from a large-scale wind farm, sampled every 4 seconds beginning 1 August 2019. With 7 397 147 observations to date, this high-frequency series captures rapid fluctuations driven by wind gusts, diurnal cycles, and turbine control strategies. The dataset, obtained from AEMO’s public data portal, has undergone standard quality checks (flagging and removal of sensor errors), resulting in a continuous series with effectively no missing entries. The dataset was aggregated from 4 seconds to 10 minutes.

The solar-power dataset \citep{solar} records power output (MW) from a utility-scale solar installation at 4-second intervals from 1 August 2019 onward (7 397 222 observations). It features pronounced diurnal patterns, cloud-induced intermittency, and seasonal tilt effects. Also sourced from AEMO, this series is complete and quality-controlled. Modeling solar generation demands that forecasting approaches address both smooth daily cycles and abrupt changes caused by passing clouds. The dataset was aggregated from 4 seconds to 10 minutes. 

\subsection{Long-term forecasting}

We evaluate all models using four metrics: MAE, RMSE, SMAPE, and MASE. These metrics jointly assess absolute error, squared error, relative percentage error, and scale-adjusted error, providing a balanced view of the forecasting performance. Each model receives a 64-step context window and predicts the subsequent 48 steps. All MoE variants use 16 experts with top-2 routing (two active experts per token). Results for the long-term forecasting setting are reported in Table 1.

\begin{table}[h!]
\centering
\caption{Performance metrics for wind, Saugeen river flow, and sunspot for long-term time-series forecasting}
\label{tab:ts_performance}
\adjustbox{width=\textwidth,center}{
\begin{tabular}{llrrrrrr}
\toprule
Dataset & Model & MAE & RMSE & SMAPE & MASE & Params \\
\midrule
{Wind} & GateTS      & \textbf{14.548} $\pm$ \textbf{0.13} & \textbf{19.798} $\pm$ \textbf{0.14} & \textbf{66.801} $\pm$ \textbf{0.64} & \textbf{0.513} & \textbf{49952} \\
       & MOEhmm      & 15.010 $\pm$ 0.12 & 20.116 $\pm$ 0.13 & 68.012 $\pm$ 0.63 & 0.529 & \textbf{49952} \\
       & MOEclassic  & 15.227 $\pm$ 0.12 & 20.259 $\pm$ 0.14 & 67.628 $\pm$ 0.66 & 0.537 & \textbf{49952} \\
       & PatchTST    & 14.876 $\pm$ 0.13 & 19.995 $\pm$ 0.14 & 66.922 $\pm$ 0.64 & 0.525 & 141200 \\
       & LSTM        & 14.915 $\pm$ 0.13 & 20.022 $\pm$ 0.14 & 66.908 $\pm$ 0.65 & 0.526 & 53552 \\
\midrule
{Saugeen} & GateTS      & \textbf{18.583} $\pm$ \textbf{0.36} & 40.659 $\pm$ 0.62 & \textbf{51.878} $\pm$ \textbf{0.34} & \textbf{0.891} & \textbf{49952} \\
          & MOEhmm      & 19.753 $\pm$ 0.34 & 40.374 $\pm$ 0.60 & 54.335 $\pm$ 0.34 & 0.947 & \textbf{49952} \\
          & MOEclassic  & 20.989 $\pm$ 0.33 & 40.812 $\pm$ 0.59 & 59.587 $\pm$ 0.36 & 1.006 & \textbf{49952} \\
          & PatchTST    & 20.152 $\pm$ 0.34 & 40.537 $\pm$ 0.59 & 55.180 $\pm$ 0.37 & 0.966 & 141200 \\
          & LSTM        & 19.329 $\pm$ 0.32 & \textbf{39.415} $\pm$ \textbf{0.58} & 53.469 $\pm$ 0.38 & 0.927 & 53552 \\
\midrule
{Sunspot} & GateTS      & \textbf{22.913} $\pm$ \textbf{0.15} & \textbf{32.994} $\pm$ \textbf{0.19} & 72.615 $\pm$ 0.60 & \textbf{0.380} & \textbf{49952} \\
          & MOEhmm      & 23.898 $\pm$ 0.15 & 33.260 $\pm$ 0.19 & 72.056 $\pm$ 0.60 & 0.396 & \textbf{49952} \\
          & MOEclassic  & 23.756 $\pm$ 0.16 & 33.326 $\pm$ 0.20 & 72.459 $\pm$ 0.62 & 0.394 & \textbf{49952} \\
          & PatchTST    & \textbf{23.054} $\pm$ \textbf{0.16} & \textbf{32.972} $\pm$ \textbf{0.20} & 73.454 $\pm$ 0.62 & \textbf{0.382} & 141200 \\
          & LSTM        & 23.507 $\pm$ 0.16 & 33.169 $\pm$ 0.19 & \textbf{71.980} $\pm$ \textbf{0.60} & 0.390 & 53552 \\
\bottomrule
\end{tabular}
}
\end{table}

Table 1 shows that GateTS attains the best overall accuracy across all evaluated forecasters. On Wind and Saugeen, it achieves the lowest errors, reducing MAE by approximately 4\% and 8–9\%, respectively, relative to the strongest baselines (PatchTST and LSTM). The advantage over the MoE variant with classic gating is especially clear: MOEclassic yields MASE > 1 (worse than a naïve forecast), whereas GateTS remains well below this threshold. On Sunspot, the top-performing models are GateTS and PatchTST. GateTS attains these gains with fewer active parameters than LSTM and roughly one-third the active parameters of PatchTST, underscoring its parameter efficiency (see the caption of Table 1 for exact counts).

\subsection{Short-term forecasting}

For short-term forecasting, the same datasets are used, but with a context size of 24 time units and a forecasting horizon of 4 time units. A mixture of experts models using 16 experts with 2 active experts. Results are shown in Table 2. 

\begin{table}[h]
\centering
\caption{Performance metrics for wind, Saugeen river flow, and sunspot for short-term time-series forecasting}
\label{tab:short_horizon_performance}
\adjustbox{width=\textwidth,center}{
\begin{tabular}{llrrrrr}
\toprule
Dataset & Model & MAE & RMSE & SMAPE & MASE & Params \\
\midrule
{Wind} 
  & GateTS      & 4.462 $\pm$ 0.07 & 7.395 $\pm$ 0.07 & 35.533 $\pm$ 0.64 & 0.139 & \textbf{45\,876} \\
  & MOEhmm      & 4.687 $\pm$ 0.07 & 7.552 $\pm$ 0.08 & 37.077 $\pm$ 0.63 & 0.146 & \textbf{45\,876} \\
  & MOEclassic  & 4.558 $\pm$ 0.07 & 7.470 $\pm$ 0.07 & 37.388 $\pm$ 0.68 & 0.142 & \textbf{45\,876} \\
  & PatchTST    & 4.500 $\pm$ 0.07 & 7.408 $\pm$ 0.08 & 36.351 $\pm$ 0.65 & 0.140 & 138\,468 \\
  & LSTM & \textbf{4.337} $\pm$ \textbf{0.07} & \textbf{7.309} $\pm$ \textbf{0.08} & \textbf{32.220} $\pm$ \textbf{0.55} & \textbf{0.135} & 50\,692 \\
\midrule
{Saugeen} 
  & GateTS     & \textbf{9.578} $\pm$ \textbf{0.45} & 27.203 $\pm$ 0.52 & \textbf{23.070} $\pm$ \textbf{0.34} & \textbf{0.476} & \textbf{45\,876} \\
  & MOEhmm      & 10.920 $\pm$ 0.46 & 28.354 $\pm$ 0.52 & 27.901 $\pm$ 0.33 & 0.542 & \textbf{45\,876} \\
  & MOEclassic  & 11.213 $\pm$ 0.46 & 27.966 $\pm$ 0.52 & 28.829 $\pm$ 0.38 & 0.557 & \textbf{45\,876} \\
  & PatchTST    & 9.779 $\pm$ 0.45 & 27.440 $\pm$ 0.51 & 23.785 $\pm$ 0.36 & 0.486 & 138\,468 \\
  & LSTM        & 9.767 $\pm$ 0.42 & \textbf{26.371} $\pm$ \textbf{0.48} & 23.982 $\pm$ 0.31 & 0.485 & 50\,692 \\
\midrule
{Sunspot} 
  & GateTS      & \textbf{14.677} $\pm$ \textbf{0.14} & \textbf{21.638} $\pm$ \textbf{0.15} & \textbf{65.290} $\pm$ \textbf{0.77} & \textbf{0.122} & \textbf{45\,876} \\
  & MOEhmm      & 14.786 $\pm$ 0.15 & 21.679 $\pm$ 0.16 & 65.815 $\pm$ 0.79 & 0.123 & \textbf{45\,876} \\
  & MOEclassic  & 15.009 $\pm$ 0.15 & 21.745 $\pm$ 0.16 & 65.797 $\pm$ 0.79 & 0.125 & \textbf{45\,876} \\
  & PatchTST & 14.730 $\pm$ 0.15 & \textbf{21.656} $\pm$ \textbf{0.16} & 66.293 $\pm$ 0.80 & 0.123 & 138\,468 \\
  & LSTM        & 14.881 $\pm$ 0.15 & 21.804 $\pm$ 0.16 & 69.965 $\pm$ 0.84 & 0.124 & 50\,692 \\
\bottomrule
\end{tabular}
}
\end{table}

Table 2 shows that, for short-horizon forecasting on the Wind series, the classical LSTM yields the lowest errors overall. Within the Mixture-of-Experts family, the proposed GateTS variant remains clearly superior, lowering MAE and RMSE by roughly 5 \% relative to the next-best MOEhmm while operating with an identical parameter budget. On the Saugeen data set, GateTS attains the strongest absolute performance and thereby improves on the two alternative MoE variants by 12–15 \% and on PatchTST by about 2 \%. On the Sunspot series, all five architectures exhibit near-identical performance, differ by less than 1\%, and their confidence intervals overlap—replicating the parity observed in the long-horizon experiment.

\subsection{Intermittent forecasting}

Intermittent forecasting tackles time series characterized by frequent, irregular stretches of zero values punctuated by occasional non-zero observations. In contrast, continuous forecasting deals with series that remain non-zero or change smoothly at every time step. This type of forecasting is represented by the solar power dataset. Context size is 64 time units, and the horizon for forecasting is 48 time units.
A mixture of expert models used 16 experts, with 2 active experts. Metrics used for comparing models were: MAE, RMSE, and MASE. SMAPE isn't helpful here because time series have a lot of zeros. Results presented in Table 3.

\begin{table}[h]
\centering
\caption{Performance of models for intermittent forecasting on the \textit{Solar}‐demand series. }
\label{tab:solar_performance}
\begin{tabular}{lrrrr}
\toprule
Model      & MAE & RMSE & MASE & Params \\
\midrule
GateTS     & \textbf{10.898} $\pm$ \textbf{0.14} & \textbf{18.838} $\pm$ \textbf{0.17} & \textbf{0.188} & \textbf{49\,952} \\
MOEhmm     & \textbf{10.875} $\pm$ \textbf{0.14} & \textbf{18.877} $\pm$ \textbf{0.17} & \textbf{0.186} & \textbf{49\,952} \\
MOEclassic & 12.775 $\pm$ 0.15 & 20.047 $\pm$ 0.17 & 0.218 & \textbf{49\,952} \\
PatchTST   & 14.068 $\pm$ 0.12 & 19.588 $\pm$ 0.13 & 0.240 & 141\,200 \\
LSTM       & 11.260 $\pm$ 0.16 & 18.936 $\pm$ 0.18 & 0.195 & 53\,552 \\
\bottomrule
\end{tabular}
\end{table}

\begin{figure}[ht]
\adjustbox{width=\textwidth,center}{\includegraphics{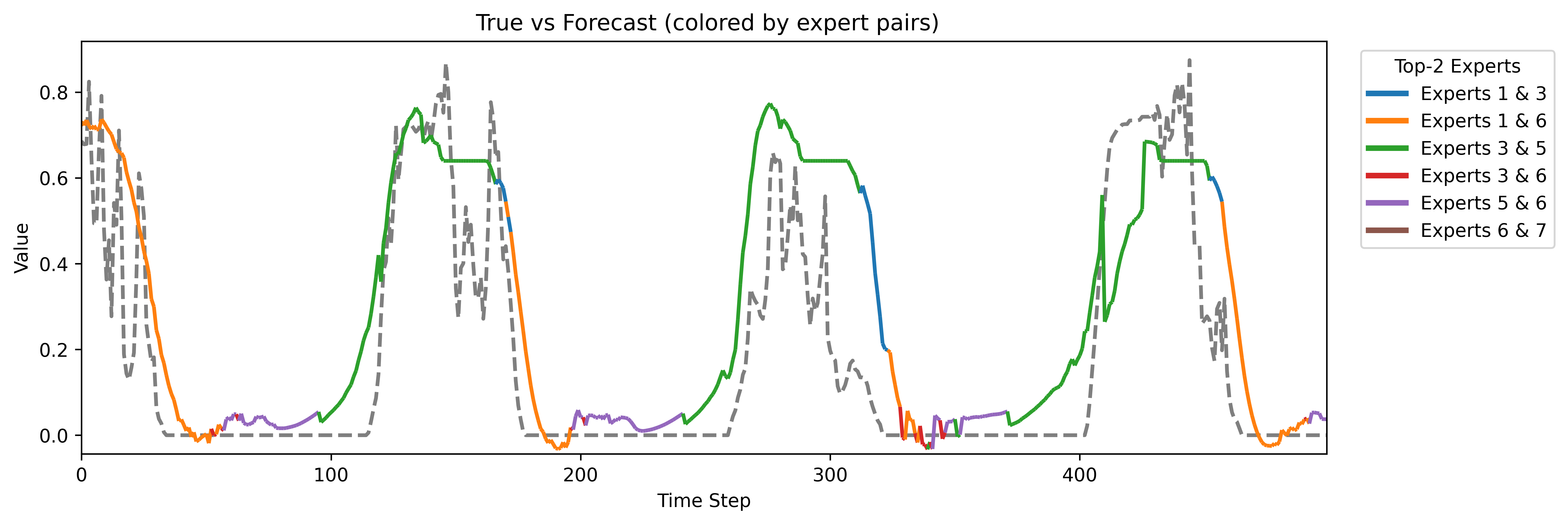}}
\caption{GateTS forecasts on the Solar Energy dataset. Colors denote the expert pair activated by the router at each time step during inference.}\label{fig:forecast}
\end{figure}

From the results, we can see that mixture of experts models with custom gating outperform other models by all metrics. Also, we can see that GateTS outperforms the mixture of experts model with classic gating by 15\% in MAE and 13\% in MASE. This happens in the proposed model; each pair of experts has become specialized in their part of the time series, as shown in Figure \ref{fig:forecast}. 

\section{Conclusion}\label{sec:conclusion}

This paper has presented a sparsely activated Mixture-of-Experts model that incorporates an attention-inspired gating mechanism specifically designed for univariate time-series forecasting, called GateTS. Recasting expert selection as a Kronecker product between input-derived keys and learnable expert queries allows the router to allocate capacity exactly where it is required, eliminating the need for auxiliary load-balancing losses and cumbersome temperature schedules. The model despite activating only a small subset of specialists per one inference step, attains or surpasses the accuracy of far larger baselines: in our experiments,  model used three times fewer active parameters than PatchTST and still delivered the best overall error scores across wind-power, Saugeen river flow, and sunspot datasets. Relative to a classical one-layer MoE gate, the proposed gating achieved lower values across all metrics on average, and it improved upon PatchTST while requiring just a fraction of its parameters. Crucially, these gains were achieved with a single end-to-end MSE objective, demonstrating that sophisticated routing behavior can emerge without additional regularisation terms or elaborate hyper-parameter tuning.

These findings indicate that a standard Mixture-of-Experts model that relies on the conventional softmax gating mechanism is sub-optimal for time-series forecasting, as the router tends to activate an excessive number of experts during inference, damaging both predictive accuracy and computational efficiency.

The results indicate that conditional computation, when paired with a carefully engineered gate, can bridge the performance gap between lightweight recurrent models and heavyweight spectral Transformers, yielding a practical and scalable solution for data-constrained forecasting problems. 

Future work should address limitations and extend the GateTS to multivariate time-series forecasting, examine its integration into large-language models to dispel auxiliary load-balancing losses during training, and probe the limits of extremely sparse routing mixtures to minimise computational overhead while maintaining predictive accuracy.

\section{Funding}
This work is supported by the European Union’s Horizon Europe research and innovation program under grant agreement No. 101138678, project ZEBAI (Innovative Methodologies for the Design of Zero-Emission and Cost-Effective Buildings Enhanced by Artificial Intelligence).

\section{Declaration of competing interest}
The authors declare that they have no known competing financial
interests or personal relationships that could have appeared to influence
the work reported in this paper.

\section{Author contributions}

The authors confirm their contribution to the paper as follows: Conceptualization: K.Y.; Data curation: K.Y.; Formal analysis: K.Y.; Funding acquisition: I.I.; Investigation: K.Y. ; Methodology: K.Y.; Project administration: K.Y., I.I. and M.L.; Resources: K.Y.;  Software: K.Y.; Supervision: I.I.; Validation: M.L., I.I.; Visualization: K.Y., M.L.; Writing – original draft: K.Y., M.L. and I.I.; Writing – review and editing: K.Y., M.L. and I.I.; All authors reviewed the results and approved the final version of the manuscript.

\section{Data availability}

Datasets are open access and can be downloaded by links presented in references \citep{solar}, \citep{sunspot}, \citep{saugeen}, \citep{wind}.

\section{Declaration of generative AI and AI-assisted technologies in the writing process}
During the preparation of this work the authors used Grammarly in order to improve the quality of the English writing. After using this tool/service, the author(s) reviewed and edited the content as needed and take(s) full responsibility for the content of the published article.

 \bibliographystyle{elsarticle-harv} 
 \bibliography{example.bib}

\end{document}